\documentclass[journal]{IEEEtran}
\usepackage{amsmath}
\usepackage{graphicx}
\usepackage{collectbox}
\usepackage{placeins}
\usepackage{multirow}
\usepackage{psfrag}
\usepackage{algorithm,algpseudocode}
\usepackage{esvect}
\usepackage{epstopdf}
\usepackage{balance}
\usepackage{cite}
\usepackage[justification=centering]{caption}
\usepackage{booktabs}

\usepackage{color}

\definecolor{Red}   {rgb}{1,0,0}

\newcommand*{\oline}[1]{%
    \kern-\fboxsep
    \vbox{%
        \hrule
        \kern2ex
        \hbox{%
            \kern\fboxsep
            #1%
            \kern\fboxsep
        }%
    }%
    \kern-\fboxsep
}
\usepackage{hyperref}
\hypersetup{
    colorlinks=true,
    linkcolor=blue,
    filecolor=magenta,
    urlcolor=cyan,
}

\begin{document}

\title{Model-Free Optimization Using Eagle Perching Optimizer}

\author{Ameer Tamoor Khan, Shuai Li \IEEEmembership{Senior Member,~IEEE}, Predrag S. Stanimirovic, Yinyan Zhang
    \thanks{This work is supported by Hong Kong Research Grants Council Early
        Career Scheme (with number 25214015), by Department General Research
        Fund of Hong Kong Polytechnic University (with number G.61.37.UA7L), and
        also by PolyU Central Research Grant (with number G-YBMU), by National
        Natural Science Foundation of China (with number 61503413) and bilateral project between
        China and Serbia  (with number 4-5).\newline
        A. Khan, S. Li, and Y. Zhang are with
        the Department of Computing, The Hong Kong Polytechnic University,
        Hung Hom, Kowloon, Hong Kong, China. Predrag S. Stanimirovic is with University of Nis, Serian.  (Corresponding author, e-mail:
        shuaili@polyu.edu.hk)}
}

\maketitle

\begin{abstract}
The paper proposes a novel nature-inspired technique of optimization. It mimics the perching nature of eagles and uses mathematical formulations to introduce a new addition to metaheuristic algorithms. The nature of the proposed algorithm is based on exploration and exploitation. The proposed algorithm is developed into two versions with some modifications. In the first phase, it undergoes a rigorous analysis to find out their performance. In the second phase it is benchmarked using ten functions of two categories; uni-modal functions and multi-modal functions. In the third phase, we conducted a detailed analysis of the algorithm by exploiting its controlling units or variables. In the fourth and last phase, we consider real world optimization problems with constraints. Both versions of the algorithm show an appreciable performance, but analysis puts more weight to the modified version. The competitive analysis shows that the proposed algorithm outperforms the other tested metaheuristic algorithms. The proposed method has better robustness and computational efficiency.
\end{abstract}


\IEEEpeerreviewmaketitle
\section{Introduction}
For the past few years, metaheuristic algorithms have played a vital role in determining the optimal solution for engineering optimization problems. They are based on stochastic approach different from the deterministic approach. In the later one can get the same solution repeatedly provided the initial conditions are the same, whereas in the former one that is not the case. Metaheuristic algorithms search randomly over the search space and every time end up at not the exact same solution as before \cite{1}. This is not a big problem for deterministic algorithms in case of uni-modal problems, which only have one global solution. On the other hand, there is a problem when we have multi-modal problems i.e. several local minima. This will create a local minima entrapment making deterministic algorithms difficult to search for the global optimum, which is certainly one of the main disadvantages of the deterministic algorithms. This is known as a local stagnation, in which an algorithm fails to find the global optimal point and stuck itself in local solutions. Real world problems are mostly like this, contains several local optima.\par
Stochastic (metaheuristic) algorithms are based on stochastic operators that ensure the randomness \cite{2}. Local stagnation or local entrapment is avoided because of the random nature and will end up at a different solution on each run despite the same initial conditions. These evolutionary algorithms at first made an initial guess by generating the random solutions called candidate solutions. The solutions are iteratively improved until the final condition is meet which is the finding of a global optimal solution. The evolutionary algorithms bring many advantages such as; simplicity, derivation independency, problem independency, and local optima avoidance \cite{3}.\par
The problem and derivation independency originate from the fact that evolutionary algorithms are based on evolutionary mechanism, they follow the certain pattern in nature which involves randomness. They treat an optimization problem as a black box and are only concerned with inputs and outputs. The main process of the optimization remains the same. They on the basis of inputs generate candidate solutions \cite{4}. To achieve the goal, they improve those candidate solutions iteratively until reaches the setted output.\par
The avoidance of local stagnation or local optima is another advantage of evolutionary algorithm, although there is no theoretical guarantee to completely avoid it, but its random nature suggests that the chances of such occurrence is narrow. The exact and accurate approximation of global optimum solution is also not guaranteed but running the algorithm for several times and then taking an average it out will improve the results.
Last but not the least, simplicity of evolutionary algorithms makes them appealing since they are based on some mechanisms that are presented in nature. The understanding of those natural phenomena’s helps us to formulate such algorithms that we can employ to solve real world problems and achieve the desired solutions. They simply follow the same rules of defined framework and treat every problem equally under the defined norms.\par
There are basically three types of optimization algorithms; basic algorithms, genetic algorithms, and swarm optimization. Basic algorithms generally involve the derivative and differential approach to solve the optimization problem. Genetic algorithms are inspired by Charles Darwin theory of evolution and is based on natural selection process that mimics the biological evolution \cite{11}. Swarm optimization is based on those species that work as a group in nature and provides each other assistance while searching food or avoiding predator; particle swarm optimization (PSO) \cite{12,13,14,15,16,17,18,19}, ant-lion (ALO) \cite{21, 22, 23, 24, 25}, and dragon optimization (DA) \cite{26, 27, 28, 29, 30} are some of those. There are some other popular algorithms which includes; grey wolf optimizer (GWO) \cite{31, 32, 33}, cuckoo optimization algorithm (COA) \cite{34, 35}, magnetic charged system search, cuckoo search (CS)
algorithm \cite{36, 37}, gravitational search algorithm (GSA) \cite{38, 39, 40, 41}, democratic particle swarm optimization (DPSO) \cite{41, 42, 43, 44}, and chaotic swarming of particles (CSP) \cite{45, 46, 47, 48, 49, 50}.\par
In this paper we propose a new algorithm called eagle perching optimizer (EPO), inspired from eagles and the way they are wired by the nature. This is another metaheuristic algorithm that iteratively for the optimal solution. The main contributions of this paper are listed as follows. \par
\begin{enumerate}
    \item The inspiration, mathematical formulation of the algorithm, global convergence proof and comparison between the designed algorithms.
    \item Results and discussion based on the comparison between EPO and other metaheuristic algorithms.
    \item Analyze the algorithm by varying its different controlling variables and will discuss the results.
    \item Engineering problems will be optimize using EPO algorithms and then we will compare the results with rest of the metaheuristic algorithm.
\end{enumerate}
The rest of this paper is organized as follows. Section \eqref{section2} presents the inspiration behind the algorithm, its mathematical formulation, EPO algorithm and its modified version, comparison between the two, and mathematical proof for global convergence. Section \eqref{section3} will discuss the benchmark testing results and performance comparisons. Section \eqref{section4} will discuss the constrained problems, in which we will optimize some real-world engineering optimization problems. Section \eqref{section5} will conclude the paper with final remarks.

\section{Eagle Perching Optimizer}\label{section2}
In this section, we will first discuss the inspiration behind the EPO algorithm. The algorithm and mathematical formulation are then discussed.

\subsection{Inspiration}
Eagle is a name generally use for many large predator birds that belong to a family of Accipitridae. They are usually 30 to 31 inches in length with the wingspan of 6 to 7 feet. They usually reside high up in the skies, even when there is a time for reproduction the male and female perform a very uncommon courtship ritual. They fly up at high altitude. There they lock their clays together and tumble down while performing aerobics move and breaks apart right before touching the ground. Female usually hatch 2 to 4 eggs, their life cycle mainly comprises of five cycles: hatching, fledgling, juvenile stage, and maturity \cite{52}.\par
\begin{figure*}[tbh]
    \centering
    \includegraphics[width=1\linewidth]{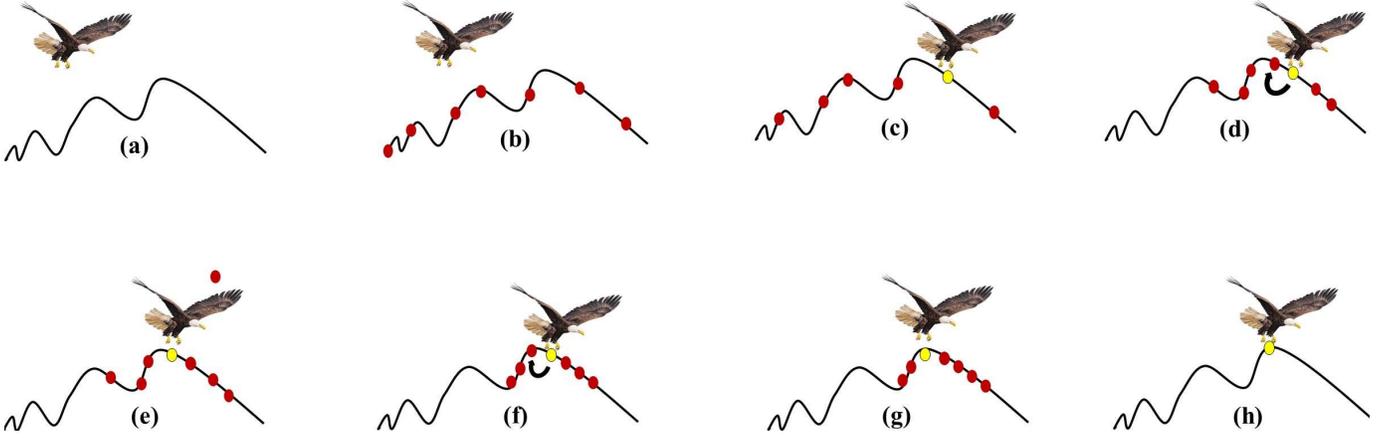}
    \caption{It shows the perching nature of the eagle, in (a) eagle is over its search space, (b) shows the sampling done by the eagle and looks for the sample at highest point, (c) shows that eagle reached over the sampled point, in (d) eagle further sampled the search space but now the space is small, it will again look for the sample at highest point, from (e) till (h) it follows the same pattern and finally reached the highest point.}
    \label{fig1}
\end{figure*}

They belong to a class of predator. They feed themselves usually on fishes, other marine life, and small animals. Their practice of hunting is unique, they fly high up in air to a possible high altitude and from there they target their prey. Once it is tracked they swoop down towards the prey and capture it. As mentioned early they reside at higher places usually on long trees, cliffs, and mountains. They have a nature gifted algorithm to track the highest place to move on, they simple alleviate high up in skies and from there they look at the ground they sample some points find the highest position out of those samples then they descend towards that point. As they come closer they again sample some points, they further clear their view regarding highest position. They iteratively execute this task and perform the fine tuning to find the highest place to reside. The nature of their built-in perching algorithm is shown in Fig. \ref{fig1}.\par

We will exploit this nature and will employ it in optimization to obtain optimal solutions. In our algorithm we will make a folk of eagles to search for the optimal elevation to reside. They all will look for the best solution individually. The algorithm afterwards will pick the best solution out of all the eagles and will compare it with previously stored best solution. This process will run iteratively until algorithm reaches its optimal solution after which no further improvement is obtained.\par
In the next section, we will discuss the mathematical representation of this algorithm that how we exploited the nature and brings it into mathematical formulation.

\subsection{Mathematical Formulation of EPO}
The EPO algorithm mimics the eagle perching behavior. Like eagle this algorithm also finds the highest point of the solution i.e. optimal solution. In optimization, there is a unique relation between the minima and maxima of a function i.e., for function $f, min(f) = max(-f)$. The nature defines the working algorithm for all its inhabitants. Eagle has a very simple but unique way to explore its terrain. Flying high up in skies it looks around by sampling few points and move towards the highest point, reaching there it again glance around and repeating the same process, this recurrence allows an eagle to reach the highest point. At first that the eagle views the whole terrain from the skies, which is exploration, after repeating the same exercise for several times it reaches near the ground, which is exploitation. The transformation from exploration to exploitation is the key for stochastic optimization (metaheuristic) algorithms. This is mathematically formulated in EPS algorithm as follows:\par
\begin{equation}\label{l.scale}
l_{scale} = l_{scale}*eta
\end{equation}

\noindent where $l_{scale}$ is the scaling variable that will decrease recurrently and will move from exploration to exploitation, $eta$ is a shrinking constant $0 < eta < 1$, which can be computed based on final value resolution.

\begin{equation}\label{eta.equ}
eta = (\frac{res}{l_{scale}})^{1/t_s}
\end{equation}
where $t_s$ is the maximum of number of iterations and $res$ is a resolution ranges with $0 < res < l_{scale}$ to restrict the $eta$ between 0 and 1. Note that if $eta > 1$ then our desired goal of exploration to exploitation will not be achieved and in each run the area of exploration will increase.\par

To achieve the optimality faster we will employ a group of i.e. eagles.These will look the search space cooperatively to make the task easier.\par
\begin{equation}\label{X.matrix}
X=
\begin{bmatrix}
X_{1,1} & X_{1,2} & X_{1,3}\hspace{0.5cm}... & X_{1,m} \\
X_{2,1} & X_{2,2} & X_{2,3}\hspace{0.5cm}... & X_{2,m} \\
X_{3,1} & X_{3,2} & X_{3,3}\hspace{0.5cm}... & X_{3,m} \\
X_{4,1} & X_{4,2} & X_{4,3}\hspace{0.5cm}... & X_{4,m} \\ \\
. & . & . & . &\vspace{-0.32cm}\\\vspace{-0.32cm}. & . & . & . &\\. & . & . & . &\\ \\
X_{n,1} & X_{n,2} & X_{n,3}\hspace{0.5cm}... & X_{n,m} \\
\end{bmatrix}
\end{equation}

\noindent where $n$ represents the number of particles that we are employing in the search space and $m$ are the number of dimensions of the search space.\par

To understand the strolling of particles (eagles) in search space, consider a particle at position $x$, to roam freely and randomly in all possible direction. A $\Delta x$ (which is a random value) is added to its current position, i.e., $x + \Delta x$ at each iteration. As a result, we have
\begin{equation}\label{X.equ}
X= X + \Delta X
\end{equation}
where,
\begin{equation}\label{delta.X}
\Delta X=
\begin{bmatrix}
R_{1,1} & R_{1,2} & R_{1,3}\hspace{0.5cm}... & R_{1,m} \\
R_{2,1} & R_{2,2} & R_{2,3}\hspace{0.5cm}... & R_{2,m} \\
R_{3,1} & R_{3,2} & R_{3,3}\hspace{0.5cm}... & R_{3,m} \\
R_{4,1} & R_{4,2} & R_{4,3}\hspace{0.5cm}... & R_{4,m} \\ \\
. & . & . & . &\vspace{-0.32cm}\\\vspace{-0.32cm}. & . & . & . &\\. & . & . & . &\\ \\
R_{n,1} & R_{n,2} & R_{n,3}\hspace{0.5cm}... & R_{n,m} \\
\end{bmatrix}
\end{equation}

\noindent with $R\in(0,1)$ denoting the random values. For each element of $X$, we have
\begin{equation}
X_{i,j} = X_{i,j} + \Delta X_{i,j}
\end{equation}
where, $i$ represents the $i_{th}$ particle and $j$ represents the $j_{th}$ dimension of the corresponding position. \\
Since the folk of eagles are high-up in air with some samples looking for the highest place on ground. They evaluate their respective set of samples and find out the highest place. We will replicate that result in our algorithm and will pass the $X_{i,j}$ to the function we want to minimize:
\begin{equation}\label{Y.equ}
Y_{i,j} = f(X_{i,j})
\end{equation}
the Equ. \eqref{Y.equ} shows that each particle position gets evaluated. Our goal is to minimize the given function so as to find the best solution out of all the eagle positions, which is denoted by $Y_{min}$. We will define two more variable $Y_{Best}$ and $X_{Best}$. The evolution of $Y_{Best}$ and $X_{Best}$ is set as follows:

\begin{align}
    \label{Y.min}
    if: Y_{min} &< Y_{Best}\\
    \label{Y.best}
    Y_{Best} &= f(X_{i,j})\\
    \label{X.best}
    X_{Best} &= X_{i,j}
\end{align}

The recursion of this algorithm will ultimately find the optimal solution of a given function.

\subsection{EPO Algorithm}
With the above explained mathematical formulation of the EPO, we are now able to discuss its algorithmic procedure in detail. The pseudo code of the EPO shown in Algorithm \eqref{Algo1}. \par
\begin{figure}\label{Algo1}
\caption{EPO algorithm}
    \begin{algorithmic}[1]
        \Procedure{}{}
        \State $initialize\ all\ the\ variables$
        \For{$<maximum\ number\ of\ iterations\>$}
        \State $calculate\ \Delta X\ using\ Equ. \eqref{delta.X}$
        \State $calculate\ X\ using\ Equ. \eqref{X.equ}$
        \For{$<total\ number\ of\ particles>$}
        \State $evaluate\ Y\ using\ Equ. \eqref{Y.equ}$
        \EndFor
        \State $evaluate\ Y_{min} from \ using\ Equ. \eqref{Y.equ}$
        \State $compare \ Y_{min}\ with\ Equ. \eqref{Y.min}$
        \If {$Equ. \eqref{Y.min}\ satisfies$}
        \State $implement \ Equ. \eqref{Y.best}\ and\ Equ. \eqref{X.best}$
        \State $reevalute\ l\_scale\ using \ Equ. \eqref{l.scale}$
        \EndIf
        \EndFor
        \EndProcedure
    \end{algorithmic}
\end{figure}


\subsection{Modified EPO Algorithm}
To accelerate the convergence of the EPO, we introduce a modification. This modification is related to the calculation of $eta$. Specifically, we will modify the value of $eta$ as well in every iteration as shown below:
\begin{equation}\label{eta.mod}
eta = eta_{max} - t* \frac{eta_{max} - eta_{min}}{t_s}
\end{equation}
where $eta_{max}$ and $eta_{min}$ represents maximum value (starting value of $eta$) and minimum value (ending value of $eta$) respectively. This will make the transformation more fast and efficient and the modified algorithm with``varying $eta$" is shown as Algorithm \eqref{Algo2}.

\begin{figure}
    \caption{Modified EPO algorithm}\label{Algo2}
    \begin{algorithmic}[1]
        \Procedure{}{}
        \State $initialize\ all\ the\ variables$
        \For{$<maximum\ number\ of\ iterations\>$}
        \State $calculate\ \Delta X\ using\ Equ. \eqref{delta.X}$
        \State $calculate\ X\ using\ Equ. \eqref{X.equ}$
        \For{$<total\ number\ of\ particles>$}
        \State $evaluate\ Y\ using\ Equ. \eqref{Y.equ}$
        \EndFor
        \State $evaluate\ Y_{min} from \ using\ Equ. \eqref{Y.equ}$
        \State $compare \ Y_{min}\ with\ Equ. \eqref{Y.min}$
        \If {$Equ. \eqref{Y.min}\ satisfies$}
        \State $implement \ Equ. \eqref{Y.best}\ and\ Equ. \eqref{X.best}$
        \State $reevalute\ l\_scale\ using \ Equ. \eqref{l.scale}$
        \State $calculate\ eta\ using \ Equ. \eqref{eta.mod}$
        \EndIf
        \EndFor
        \EndProcedure
    \end{algorithmic}
\end{figure}

The algorithm of this modified version is same as that of ``EPO Algorithm" except that in the current version we need to update the value of $eta$ as well. In the next section we will compare both these algorithms to find out which one is more efficient and accurate to find the optimality of test functions.\par
We introduced another improvement to the algorithm. 
After evaluating the values of all the particles using Equ. \eqref{Y.equ}, we sort them from best solution to the worst Equ. \eqref{Y.sort1}. Out of the sorted solutions we select ``$n$" number of best solutions and store their corresponding coordinates in an array Equ. \eqref{X.sort}. We average out the array as shown in Equ. \eqref{X.avg} and the resultant is use to again calculate the value of function using Equ. \eqref{Y.equ}. This procedure repeats recursively until an optimal solution is achieved. The purpose of this improvement is to further broaden the scope of algorithm so instead of relying on single $X_{best}$ value now we are relying on an average value $X_{avg}$.

\begin{align}
\label{Y.sort1}
Y_{sort} &= [Y_{best_1}\ Y_{best_2}\ Y_{best_3}\ Y_{best_4}\ ...\ Y_{wrost} ]\\
\label{Y.sort2}
Y_{sort} &= [Y_{best_1}\ Y_{best_2}\ Y_{best_3}\ Y_{best_4}\ ...\ Y_{best_n} ]\\
\label{X.sort}
X_{sort} &= [X_{best_1}\ X_{best_2}\ X_{best_3}\ X_{best_4}\ ...\ X_{best_n} ]\\
\label{X.avg}
X_{avg} &= \frac{X_{best_1}\ + X_{best_2}\ + X_{best_3}\ ...\ + X_{best_n}}{n}
\end{align}

\subsection{Comparison between the two algorithms}\label{Comp.EPO}
Our main goal is to optimize the function and search for the best possible solution. The general framework of our algorithm is to employ all the particles in search space and then look for the one at best position. This iterative process will finally converge the function to its optimal value. Both the algorithms perform the same job and we test some uni-modal and multi-modal functions to evaluate their performance and efficiency. For the test purposes we ran both the algorithms 30 time with 500 iterations with $l\_scale = 500$ and $res = 0.05$. For the modified EPO the value of $eta$ deteriorated from 0.9 and 0.8.

\subsubsection{Comparison based on uni-modal functions} The functions mentioned in Table \eqref{uni_EPO} are uni-modal since they have only one convergence point \cite{57}, all of which have $0$ as an optimal solution. There corresponding results are shown in Table \eqref{uni_EPO_results} from which it is evident that the modified EPO out-shined the EPO. The $avg$ of the modified EPO is more closer to actual convergence point, with small $std$, which is mainly because of the linear transformation between the exploration and the exploitation.

\begin{table}[h]
    \caption{ Uni-modal functions}
    \label{uni_EPO}
    \begin{tabular} {l    l l l}
        \hline
        Function\hspace{1cm}    &Dim    &Range  &$f_{min}$ \\
        \hline \\
        $g_1(x) = \Sigma_{i = 1}^n (x+2)_i^2 +2$\hspace{1cm}    &$30$       &$[-10\hspace{0.2cm} 10]$   &$2$ \\
        $g_2(x) = \Sigma_{i = 1}^n |x_i^2| + \Pi_{i = 1}^n |x_i|$\hspace{1cm}   &$30$       &$[-10\hspace{0.2cm} 10]$   &$0$ \\
        $g_3(x) = \Sigma_{i = 1}^n (\Sigma_{j-1}^i x_j)^2$\hspace{1cm}  &$30$       &$[-10\hspace{0.2cm} 10]$   &$0$ \\
        \hline
    \end{tabular}
\end{table}

\begin{table*}[h]
    \centering
    \caption{Results of uni-modal functions}
    \label{uni_EPO_results}
    \begin{tabular} {c c c c c c c c c c c c c}
        \hline
        Function \hspace{0.6cm} & & &$x_1$  &$x_2$ &$x_3$ &$x_4$ &$x_5$ &$f_{x}$ \\
        \hline \\
        \multirow{4}{*}{$g_1(x)$}\hspace{0.6cm}
        &\multirow{2}{*}{EPO} \hspace{0.5cm}
        &avg &-2.00E+00&-2.00E+00&-2.00E+00&-2.00E+00&-1.98E+00&-2.01E+00&-2.01E+00
        &2.0116\\
        & &std &2.570E-03&0.0101586&0.0064327&1.120E-02&0.016015&0.010055&0.018598
        &0.00168
        \\
        &\multirow{2}{*}{EPO (mod.)}\hspace{0.5cm}
        &avg &-2.00E+00&-2.00E+00&-2.00E+00&-2.00E+00&-2.00E+00&-2.00E+00&-2.00E+00
        &2\\
        & &std &2.34E-16&4.68E-16&0.00E+00&2.34E-16&0.00E+00&0.00E+00&0.00E+00
        &0
        \\ \\
        \multirow{4}{*}{$g_2(x)$}\hspace{0.6cm}
        &\multirow{2}{*}{EPO} \hspace{0.5cm}
        &avg &3.70E-03&-1.33E-02&2.69E-02&-5.08E-03&1.18E-04&9.50E-03&4.12E-03
        &0.4388\\
        & &std &1.01E-02&0.009760&0.008551&3.910E-03&0.018265&0.012637&0.020819
        &0.0604
        \\
        &\multirow{2}{*}{EPO (mod.)}\hspace{0.5cm}
        &avg &-4.18E-24&-8.58E-24&5.51E-24&3.26E-24&1.95E-23&2.38E-24&-1.81E-23
        &2.934E-22  \\
        & &std &7.74E-40&1.55E-39&0.00E+00&7.74E-40&0.00E+00&3.87E-40&3.10E-39
        &4.956E-38
        \\ \\
        \multirow{4}{*}{$g_3(x)$}\hspace{0.6cm}
        &\multirow{2}{*}{EPO} \hspace{0.5cm}
        &avg &1.78E-02&-1.93E-02&4.19E-03&2.43E-02&-1.30E-02&-6.35E-03&-3.04E-02
        &0.0364\\
        & &std &1.94E-02&0.012850&0.008921&4.59E-03&0.0155342&0.016608&0.012380
        &0.00284
        \\
        &\multirow{2}{*}{EPO (mod.)}\hspace{0.5cm}
        &avg &-5.53E-24&2.17E-23&-4.43E-24&-1.26E-23&-1.28E-23&-1.95E-23&-1.24E-23
        &2.189E-23  \\
        & &std &7.74E-40&0.00E+00&0.00E+00&0.00E+00&0.00E+00&3.10E-39&3.10E-39
        &0.00E+00
        \\ \\
        \hline

    \end{tabular}
\end{table*}

From Table \eqref{uni_EPO} it is quite evident that the modified EPO algorithm has more promising results than the EPO with constant $eta$. That is only because in the modified version we have better control over the scaling of the search space.\\
We used the built-in MATLAB command \texttt{fmincon} to evaluate the optimal solution of all functions mentioned, and then compared those results with both the algorithms. Both were efficient and approximated the optimal solution, but the meticulous comparison showed that the modified EPO outclassed the simple EPO. In the modified EPO algorithm we iteratively change the value of $eta$ from 0.9 to 0.8, since all uni-modal function converges to $0$ so the amount of accuracy in case of the modified EPO was between range $10^{-20}$ and $10^{-50}$ with standard deviation(std) between the range $10^{-20}$ and $10^{-60}$  whereas the EPO with constant $eta$ produced results with accuracy in range $10^{-1}$ and $10^{-2}$ with standard deviation (std) between $10^{-1}$ and $10^{-3}$.\\
\subsubsection{Comparison based on multi-modal functions}
We applied the algorithms on multi-modal functions as well. In such functions, there are many local optimum solutions that lie at different locations within the search space but there is only one optimal or global solution \cite{61}. Our algorithm is meta heuristic based, so there will be a randomness in our solution despite of same initial conditions. Applying it on multi-modal functions can produce different results, which is why we ran it for 30 times and calculated the standard deviation to finds out the how consistent and accurate our algorithm is. The functions are shown in the Table \eqref{multi_EPO} and their respective results are shown in Table \eqref{multi_EPO_results}.

\begin{table*}[h]
    \begin{center}
        \caption{ Multi-modal functions}
        \label{multi_EPO}
        \begin{tabular} {l    l l l}
            \hline
            Function\hspace{1cm}    &Dim    &Range  &$f_{min}$ \\
            \hline \\
            $g_4(x) = -20exp(-0.2\sqrt{\frac{1}{n}}\sigma_{i=n}^nx_i^2)-exp(\frac{1}{n}\sigma_{i=n}^n\cos(2\pi x_i))+20+e$ \hspace{3cm} &$30$       &$[-5.12\hspace{0.2cm} 5.12]$   &$0$ \\
            $g_5(x) = (1(1+x_1+x_2)^2*(19-14x_1+3*x_1^2-14*x_2+6*x_1*x_2+3*x_2^2))*$ \hspace{3cm}   &$30$       &$[-10\hspace{0.2cm} 10]$   &$3$  \\  \hspace{1.15cm}$(30+(2*x_1-3*x_2^2*18-32*x_1=12*x_1^2+48*x_2-36*x_1*x_2+27*x_2^2)))$ & & \\
            $g_6(x) = \frac{1}{4000}\Sigma_{i = 1}^n x_i^2 - \prod_i^n \cos(\frac{x_i}{i}) + 1$ \hspace{3cm}    &$30$       &$[-10\hspace{0.2cm} 10]$   &$0$ \\ \\
            \hline
        \end{tabular}
    \end{center}
\end{table*}

\begin{table*}[h]
    \centering
    \caption{Results of multi-modal functions}
    \label{multi_EPO_results}
    \begin{tabular} {c c c c c c c c c c c c c c c}
        \hline
        Function \hspace{0.6cm} & & &$x_1$  &$x_2$ &$x_3$ &$x_4$ &$x_5$ &$x_6$ &$x_7$   &$f_{x}$ \\
        \hline \\
        \multirow{4}{*}{$g_4(x)$}\hspace{0.6cm}
        &\multirow{2}{*}{EPO} \hspace{0.5cm}
        &avg &-4.21E-03&-4.83E-03&-1.08E-02&-1.08E-02&1.99E-02&-1.62E-03&2.67E-02&-8.713E+12\\
        & &std &1.68E-02&0.0118578&0.0107313&1.30E-02&1.46E-02&7.70E-03&1.67E-02
        &3.247E+11
        \\
        &\multirow{2}{*}{EPO (mod.)} \hspace{0.5cm}
        &avg &1.03E+03&1.03E+03&1.03E+03&1.03E+03&1.03E+03&1.03E+03&1.03E+03
        &-1.068E+13\\
        & &std &2.40E-13&2.40E-13&0.00E+00&0.00E+00&2.40E-13&2.40E-13&2.40E-13
        &0.0020
        \\ \\
        \multirow{4}{*}{$g_5(x)$}\hspace{0.6cm}
        &\multirow{2}{*}{EPO} \hspace{0.5cm}
        &avg &2.38E-02&-9.96E-01&-1.04E-01&-6.82E-01&4.70E-02&-3.56E-01&-6.62E-01
        &3.1304\\
        & &std &0.00E+00&2.34E-16&0.00E+00&0.00E+00&0.00E+00&0.00E+00&1.17E-16
        &4.681e-16
        \\
        &\multirow{2}{*}{EPO (mod.)}\hspace{0.5cm}
        &avg &-1.95E-03&-9.91E-01&-9.92E-01&-1.97E-01&-7.14E-01&-4.45E-01&1.63E-01
        &3.0376\\
        & &std &0.00E+00&1.17E-16&1.17E-16&2.93E-17&1.17E-16&0.00E+00&2.93E-17
        &0.00E+00
        \\ \\
        \multirow{4}{*}{$g_6(x)$}\hspace{0.6cm}
        &\multirow{2}{*}{EPO} \hspace{0.5cm}
        &avg &-5.53E-24&2.17E-23&-4.43E-24&-1.26E-23&-1.28E-23&3.21E-03&-1.29E-02
        &0.00E+00\\
        & &std &7.74E-40&0.00E+00&0.00E+00&0.00E+00&0.00E+00&1.03E-02&1.83E-02
        &0.00E+00
        \\
        &\multirow{2}{*}{EPO (mod.)} \hspace{0.5cm}
        &avg &4.68E-10&9.20E-04&8.38E-03&2.90E-03&1.08E-02&-1.92E-09&-6.14E-09
        &0.005720\\
        & &std &1.56E-02&2.66E-02&0.01653&9.34E-03&1.78E-02&0.00E+00&8.72E-25
        &0.000947
        \\ \\
        \hline
    \end{tabular}
\end{table*}

\begin{table*}[h]
    \centering
    \caption{Results of further modified EPO algorithm}
    \label{EPO-mod-mod}
    \begin{tabular} {llllllllllllll}
        \hline
        Function  &  &$n = 2$   &$n = 3$ &$n = 4$ &$n = 5$ &$n = 6$ &$n = 7$ &$n = 8$ &$n = 9$ &$n = 10$ &$n = 11$ &$n = 12$  \\
        \hline \\
        $g_2(x)$
        &avg &1.82E-24&7.77E-24&4.08E-24&2.54E-18&2.18E-19&4.63E-16&6.08E-08&2.92E-05
        &3.74E-04&5.14&0.0035\\
        $g_3(x)$
        &avg &1.75E-42&3.05E-37&5.09E-22&3.50E-14&5.39E-08&2.96E-13&1.13E-05&8.13E-06&0.675710098&0.15&4.73\\
        $g_4(x)$
        &avg &5.71&5.62&7.518&7.08&6.29&8.74&8.76&8.39&7.82&7.30&8.45\\
        $g_5(x)$
        &avg &0.068&0.039&0.056&0.051&0.027&0.209&0.101&0.108&0.194&0.209&0.402\\ \\
        \hline
    \end{tabular}
\end{table*}

The results presented in Table \eqref{multi-modal-algos-results} further make it clear that the modified EPO is more efficient than EPO with constant $eta$. As mentioned above multi-modal functions have several optimal solutions within the search space so it may be a difficult task for an algorithm especially metaheuristic one to identify the most accurate optimal solution. Considering this fact, we defined a parameter ``std" to finds out the standard deviation that occurs in 30 runs. It is obvious from the result that ``std" in case of the modified EPO is of $10^{-2}$ value which is comparable with the performance of EPO accuracy, still the modified EPO has shown more efficient and consistent results.\par
We did the analysis of improved version of the algorithm as well, mentioned from Equ. \eqref{Y.sort1} to Equ. \eqref{X.avg} and found that as the value of $n$ increases the efficiency of the algorithm decreases, we applied the algorithm on both uni-modal and multi-modal functions and found out the results, shown in Table \eqref{EPO-mod-mod}.
The parameters kept were: $t_{s(iterations)} =500$, $res =0.05$, $dim = 4$, $k_{particles} =30$, and $eta$ varies from $0.9$ to $0.8$.\par
In the next section, we will verify that using EPO the convergence of a function to its optimality is inevitable.

\subsection{Mathematical proof for global convergence}
A series converges when sequence of its partial sums approaches to a limit; that means, the partial sums become closer to a given number when the number of terms increase. In a proof we will show that the convergence of EPO is inevitable, the solution converges to global optima with probability 1 when time $t$ goes to infinity. Before starting the mathematical manipulations, we need to consider these three points.
\begin{enumerate}
    \item  Monotonically  non-reducing function converges if it has an
    upper bound.
    \item The (1) does not imply that the lower bound is the limit.
    The limit may be greater than it.
    \item Here convergence means the limit exists. It does not imply the limit is exactly the global optima.
\end{enumerate}

Monotonicity recall Equ. \eqref{Y.min} Equ. \eqref{Y.best} and Equ. \eqref{X.best}, with this we know $Y_{Best}(t+1) \geq Y_{Best}(t)$. if condition Equ. \eqref{Y.min}  is not met then, $Y_{Best}(t+1) = Y_{Best}(t)$. For upper bound recall that we are searching for the maximum.  If the maximum exists, itself constructs the upper bound.\par

This proof is particularly for proving that the convergence limit is exactly the global optima. In order to proceed let's do a little modification to the Equ. \eqref{l.scale} as show below.
\begin{equation}\label{l.scale.mod}
l\_scale = l\_scale * eta + L_o \hspace{0.5cm} 0 < L_o < 1
\end{equation}

The equations Equ. \eqref{X.equ} and Equ. \eqref{l.scale.mod} plays the key role in this proof. Consider the set of global optima $x_{opt}$ as follow:
\begin{equation}\label{X.opt}
{x\in R^n,  abs(x_i-x_opt_i)\leq delta}
\end{equation}
where $delta > 0  and x = [x_i]$.\\
\textbf{Consider the problem:}  for step t, how large is the probability to cast a sample into the above defined neighborhood?

\begin{align}
    \begin{split}
        \Omega = \{x\in R^n, x_opt_i-delta<x_i\\
        +randn<x_opt_i+delta\}
    \end{split}
\end{align}

Thus the above Probability will be: \par

\begin{align}
    \begin{split}
        P &= P(x_{opt\_1}-delta<x_1+randn_1\\
        &<x_{opt\_1}+delta)*P(x_{opt\_2}-delta\\
        &<x_2+randn_2<x_{opt\_2}+delta)\\
        &*...*P(x_{opt\_n}-delta<x_n+randn_n\\
        &<x_{opt\_n}+delta)\\
        P &= [F(x_{opt\_1}+delta-x_1)-F(x_{opt\_1}\\
        &-delta-x_1)]*[F(x_{opt\_2}+delta-x_2)\\
        &-F(x_{opt\_2}-delta-x_2)]*...*[F(x_{opt\_n}\\
        &+delta-x_n)-F(x_{opt\_n}-delta-x_n)]\\
    \end{split}
\end{align}
where $F(x)$ is the cdf (cumulative distribution function) of $N(0,L)$, i.e., $F(x) = int\_(-inf)^x*f(y)dy$.\\
We made the assumption that searching space is bounded. And the diameter
of searching space, which is defined as the largest distance between any two points in the domain, is bounded by D. Then, we can conclude that,\\
\begin{equation}
\begin{split}
[F(x+delta)-F(x-delta)]&>min\{f(x+delta)\\
&,f(x-delta)\}\\
&*2*delta\\
\end{split}
\end{equation}
For, $[F(x_{opt\_i}+delta-x_i)-F(x_{opt\_i}-delta-x_i)]$, note that
$abs(x_{opt\_i}-x_i)\leq D$ according to above definition. Thus, \\
\begin{equation}
\begin{split}
&[F(x_{opt\_i}+delta-x_i)-F(x_{opt\_i}-delta-x_i)]>\\
&min\{f(D+delta),f(D-delta)\}*2*delta
\end{split}
\end{equation}
where $f(x)$ is the pdf of $N(0,1)$ Clearly, $f(D+delta)<f(D-delta)$ if $delta<<D$.
So, $min=f(D+delta)$.Therefore, $[F(x_{opt\_i}+delta-x_i)-F(x_{opt\_i}-delta-x_i)] >f(D+delta)*2*delta$, for $N(0,1)$ $f(D+delta)$
Can be computed easily. From the above we know the probability of $Y \geq [f(D+delta)*2*delta]^n=epsilon>0$ where $epsilon > 0$. In order to understand how proof works consider the descriptive form below.
\begin{itemize}
    \item Consider $K$ particles, and their probability of falling in the neighborhood of optimal solution is $epsilon$.
    \item The probability of $K$ particles never falling in the neighborhood of optimal solution is $(1-epsilon)^k$.
    \item For $t$ number of iterations this probability will be:\\
    $(1-epsilon)^k*...(1-epsilon)^k=(1- epsilon)^{kt}$\\
    This represent the probability that sample will never drop into the optima neighborhood. So far, we have successfully proved that at step $t$, the probability to sample a point in the neighborhood of the optimum is greater than $epsilon$.
    \item When $t$ goes to infinity, $(1-epsilon)^{kt}$ goes to $0$. This means that the probability of $K$ particles not falling in the neighborhood of optimal solution reaches to zero as the $t$ approaches to infinity. Thus the probability of particles falling in neighborhood of optimal solution becomes 1.
\end{itemize}

\section{Results and discussion}\label{section3}
In this section, we will do the competitive analysis and will benchmark the performance of EPO with many other famous metaheuristic algorithms. We will use number of test functions, based on their nature they are divided into two groups. uni-modal functions and multi-modal functions. Uni-modal functions are those with single optimum solution, so they are easy to handle since function converges to a single solution. Contrary, multi-modal functions are those with number of optimal solution although they also have one global convergence but because of the presence of number of optimal solutions it is hard to deal with them. \par
For the results verification of EPO, we will compare the results with ant-lion optimizer (ALO), dragon fly optimizer (DA), particle swarm optimizer (PSO) which is the best among the group of swarm optimizers, and GA the best evolution-based optimization algorithm. In addition to them, there are other recently developed optimization techniques which includes; flower pollination algorithm, state of matter search algorithm (SMS), cuckoo search algorithm (CS), bat algorithm (BA), and firefly algorithm (FA). In order to quantify the results, we will run each function 30 times and will calculate their average (avg) and standard deviation (std).\par
Each of the function will undergo 30 test run, with 500 iteration, $l\_scale = 500$, and $res = 0.05$. Here we will employ both the EPO algorithm and will see their comparison with the rest of the algorithms, mentioned above.

\subsubsection{Comparison result of uni-modal functions}

\begin{figure*}[]
    \centering
    \includegraphics[width=1\linewidth]{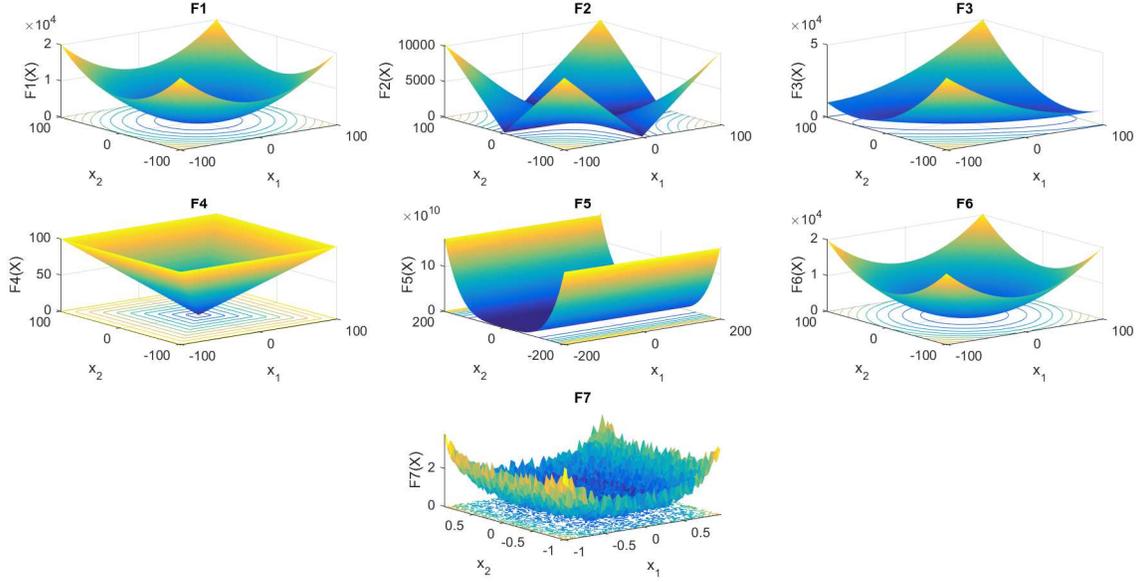}
    \caption{It represents the search space of all the seven functions from $F1$ to $F7$ of the uni-modal nature. They all have only one global minima and in this case its zero ($0$).}
    \label{SP1}
\end{figure*}

The Table \eqref{uni-modal-algos-results} shows all the seven test functions used for the comparison, the search space of these functions is also shown in Fig. \ref{SP1}. The Table \eqref{uni-modal-algos-results} shows results obtained after testing the uni-modal functions and it is quite prominent that EPO outperform rest of the algorithms. As mentioned in the section \eqref{Comp.EPO} the range of accuracy both the EPO algorithm shows is incredible. All the test functions in uni-modal has a global optimum at $0$ (zero).\\
\begin{table}[]
    \caption{ Uni-modal benchmark functions}
    \label{uni-modal-algos}
    \begin{tabular} {l    l l l}
        \hline
        Function\hspace{1cm}    &Dim    &Range  &$f_{min}$ \\
        \hline \\
        $F_1(x) = \Sigma_{i = 1}^n x_i^2$   &$30$       &$[-100, 100]$  &$0$ \\

        $F_2(x) = \Sigma_{i = 1}^n |x_i^2| + \Pi_{i = 1}^n |x_i|$   &$30$       &$[-10,10]$ &$0$ \\

        $F_3(x) = \Sigma_{i = 1}^n (\Sigma_{j-1}^i x_j)^2$  &$30$       &$[-100, 100]$  &$0$ \\

        $F_4(x) = max (|x|)$\hspace{0.5cm} $1\leq i \leq n$     &$30$       &$[-100, 100]$  &$0$ \\

        $F_5(x) = \Sigma_{i = 1}^n[100(x_{i+1} - x_i^2)^2$  &$30$       &$[-30, 30]$    &$0$ \\
        \hspace{1.1cm}$+(x_i -1)^2]$ \\

        $F_6(x) = \Sigma_{i=1}^n([x_i + 0.5])^2$    &$30$       &$[-100, 100]$  &$0$\\

        $F_7(x) = \Sigma_{i=1}^n ix_i^4 + random[0,1]$  &$30$       &$[-1.28, 1.28]$    &$0$\\
        \hline
    \end{tabular}
\end{table}

\begin{table*}
    \begin{center}
        \caption{ Results of uni-modal benchmark functions}
        \label{uni-modal-algos-results}
        \begin{tabular} {c c c c c c c c c c c c c c c}
            \hline \\
            \multirow{1}{*}{Function}
            &\multicolumn{2}{l}{EPO (mod.)}  &\multicolumn{1}{c}{}
            &\multicolumn{2}{l}{EPO}         &\multicolumn{1}{c}{}
            &\multicolumn{2}{l}{DA}          &\multicolumn{1}{c}{}
            &\multicolumn{2}{l}{ALO}         &\multicolumn{1}{c}{}
            &\multicolumn{2}{l}{PSO} \\
            \cline{2-3} \cline{5-6} \cline{8-9} \cline{11-12} \cline{14-15}  \\
            & avg &std &
            & avg &std &
            & avg &std &
            & avg &std &
            & avg &std\\
            \hline \\
            F1
            &3.93E-45 &6.56E-61 & 
            &1.14E-02 &1.66E-03 & 
            &2.85E-18 &7.16E-18 & 
            &2.59E-10 &1.65E-10 & 
            &2.70E-09 &1.00E-09\\ 
            F2
            &3.26E-22 &4.96E-38 & 
            &4.39E-01 &5.83E-02 & 
            &1.49E-05 &3.76E-05 & 
            &1.84E-06 &6.58E-07 & 
            &7.15E-05 &2.26E-05\\ 
            F3
            &1.04E-52 &1.95E-68 & 
            &8.47E-09 &2.45E-08 & 
            &1.29E-06 &2.10E-06 & 
            &6.06E-10 &6.34E-10 & 
            &4.71E-06 &1.49E-06\\ 
            F4
            &1.04E-52 &1.95E-68 & 
            &8.47E-09 &2.45E-08 & 
            &1.29E-06 &2.10E-06 & 
            &6.06E-10 &6.34E-10 & 
            &4.71E-06 &1.49E-06\\ 
            F5
            &0.00E+00 &0.00E+00 & 
            &0.00E+00 &0.00E+00 & 
            &7.600558 &6.786473 & 
            &0.346772 &0.109584 & 
            &0.123401 &2.16E-01\\ 
            F6
            &0.00E+00 &0.00E+00 & 
            &8.03E-03 &3.30E-03 & 
            &4.17E-16 &1.32E-01 & 
            &2.53E-10 &1.09E-10 & 
            &5.23E-07 &2.74E-06\\ 
            F7
            &0.001776 &0.00E+00 & 
            &0.001388 &2.64E-03 & 
            &0.010293 &4.69E-03 & 
            &0.004292 &0.005089 & 
            &0.001398 &0.001269\\ \\ 

            \multirow{1}{*}{}         \hspace{1cm}
            &\multicolumn{2}{l}{SMS}  \hspace{1cm} &\multicolumn{1}{c}{}
            &\multicolumn{2}{l}{BA}   \hspace{1cm} &\multicolumn{1}{c}{}
            &\multicolumn{2}{l}{FPA}  \hspace{1cm} &\multicolumn{1}{c}{}
            &\multicolumn{2}{l}{CS}   \hspace{1cm} &\multicolumn{1}{c}{}
            &\multicolumn{2}{l}{FA} \\
            \cline{2-3} \cline{5-6} \cline{8-9} \cline{11-12} \cline{14-15}  \\
            & avg &std &
            & avg &std &
            & avg &std &
            & avg &std &
            & avg &std\\
            \hline \\
            F1
            &0.056987 &0.014689 & 
            &0.773622 &0.528134 & 
            &1.06E-07 &1.27E-07 & 
            &6.50E-03 &2.05E-04 & 
            &0.039615 &0.01449\\ 
            F2
            &0.006848 &0.001577 & 
            &0.334583 &3.816022 & 
            &0.000624 &0.000176 & 
            &2.12E-01 &3.98E-02 & 
            &0.050346 &0.012348\\ 
            F3
            &9.60E-01 &8.23E-01 & 
            &1.15E-01 &0.766036 & 
            &5.67E-08 &3.90E-08 & 
            &2.47E-01 &2.14E-02 & 
            &4.93E-02 &0.019409\\ 
            F4
            &2.77E-01 &5.74E-03 & 
            &1.92E-01 &0.890266 & 
            &0.003837 &0.002186 & 
            &1.12E-05 &8.25E-06 & 
            &0.145513 &0.031171\\ 
            F5
            &0.085348 &0.140149 & 
            &0.334077 &0.300037 & 
            &0.781200 &0.366891 & 
            &0.007197 &7.22E-03 & 
            &2.175892 &1.447251\\ 
            F6
            &0.125323 &0.084998 & 
            &0.778849 &0.673920 & 
            &1.09E-07 &1.25E-07 & 
            &5.95E-05 &1.08E-06 & 
            &0.05873 &0.014477\\ 
            F7
            &0.000304 &0.000258 & 
            &0.137483 &0.112671 & 
            &0.003105 &0.001367 & 
            &0.001321 &0.000728 & 
            &0.000853 &0.000504\\ 
            \hline
        \end{tabular}
    \end{center}
\end{table*}
\begin{table*}[]
    \caption{ Multi-modal benchmark functions}
    \label{multi-modal-algos}
    \begin{tabular} {l    l l l}
        \hline
        Function\hspace{1cm}    &Dim    &Range  &$f_{min}$ \\
        \hline \\
        $F_8(x) = \Sigma_{i = 1}^n -x_i\sin(\sqrt{|x_i|})$  &$30$       &$[-500, 500]$  &$-418.9829xDim^a$ \\

        $F_9(x) = \Sigma_{i = 1}^n [x_i^2 -10\cos(2\pi x_i) + 10]$  &$30$       &$[-5.12,5.12]$ &$0$ \\

        $F_{10}(x) =-20exp(-0.2\sqrt{\frac{1}{n}}\sigma_{i=n}^nx_i^2)
        -exp(\frac{1}{n}\sigma_{i=n}^n\cos(2\pi x_i))+20+e$ \hspace{3cm}    &$30$&$[-5.12\hspace{0.2cm} 5.12]$  &$0$ \\
        \hline
    \end{tabular}
\end{table*}

\begin{table*}[]
    \begin{center}
        \caption{ Results of multi-modal benchmark functions}
        \label{multi-modal-algos-results}
        \begin{tabular} {c c c c c c c c c c c c c c c}
            \hline \\
            \multirow{1}{*}{Function}
            &\multicolumn{2}{l}{EPO (modified)}   &\multicolumn{1}{c}{}  &\multicolumn{2}{l}{EPO}              &\multicolumn{1}{c}{}
            &\multicolumn{2}{l}{DA}               &\multicolumn{1}{c}{}
            &\multicolumn{2}{l}{ALO}              &\multicolumn{1}{c}{}
            &\multicolumn{2}{l}{PSO} \\
            \cline{2-3} \cline{5-6} \cline{8-9} \cline{11-12} \cline{14-15}  \\
            & avg &std &
            & avg &std &
            & avg &std &
            & avg &std &
            & avg &std\\
            \hline \\
            F8
            &-62312.48 &7.67E-12 & 
            &-71195.12 &0.00E+00 & 
            &-2857.58 &383.6466 & 
            &-1606.243 &314.4302 & 
            &-1367.01 &146.4089\\ 
            F9
            &-290.000 &0.00E+00 & 
            &-286.160 &5.99E-14 & 
            &16.01883 &9.479113 & 
            &7.71E-06 &8.45E-06 & 
            &0.278588 &0.218991\\ 
            \hspace{0.15cm}F10
            &-1.068E-13 &0.002058 & 
            &-8.07E-12 &1.4709E-11 & 
            &0.23103 &0.487053 & 
            &3.73E-15 &1.50E-15 & 
            &1.11E-09 &2.39E-11\\ \\ 

            \multirow{1}{*}{}         \hspace{1cm}
            &\multicolumn{2}{l}{SMS}  \hspace{1cm} &\multicolumn{1}{c}{}
            &\multicolumn{2}{l}{BA}   \hspace{1cm} &\multicolumn{1}{c}{}
            &\multicolumn{2}{l}{FPA}  \hspace{1cm} &\multicolumn{1}{c}{}
            &\multicolumn{2}{l}{CS}   \hspace{1cm} &\multicolumn{1}{c}{}
            &\multicolumn{2}{l}{FA} \\
            \cline{2-3} \cline{5-6} \cline{8-9} \cline{11-12} \cline{14-15}  \\
            & avg &std &
            & avg &std &
            & avg &std &
            & avg &std &
            & avg &std\\
            \hline \\
            F8
            &-4.20735 &9.36E-16 & 
            &-1065.88 &858.498 & 
            &-1842.42 &50.42824 & 
            &-2094.91 &0.007616 & 
            &-1245.59 &353.2667\\ 
            F9
            &1.32512 &0.326239 & 
            &1.233748 &0.686447 & 
            &0.273294 &0.068583 & 
            &0.127328 &0.002655 & 
            &0.263458 &0.182824\\ 
            \hspace{0.15cm}F10
            &8.88E-06 &8.56E-09 & 
            &0.129359 &0.043251 & 
            &0.007398 &0.007096 & 
            &8.16E-09 &1.63E-08 & 
            &0.168306 &0.050796\\ 
        \end{tabular}
    \end{center}
\end{table*}

\begin{figure*}[]
    \centering
    \includegraphics[width=1\linewidth]{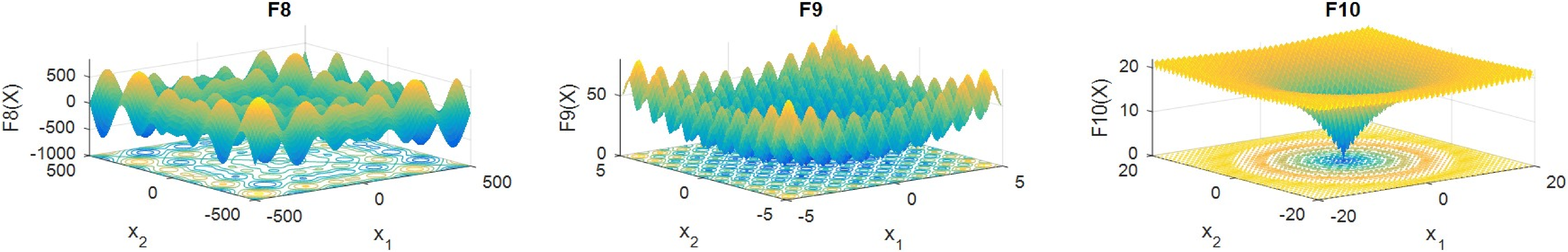}
    \caption{It represents the search space of all three functions from $F8$ to $F10$ of the multi-modal nature. They all have more than one local-minima but one global minima, in case of $F8$ it depends on the dimensions of the algorithms otherwise rest of the two has global minimum at zero ($0$).}
    \label{SP2}
\end{figure*}

\subsubsection{Comparison result of multi-modal functions}
The comparison results of the multi-modal functions are shown in Table \eqref{multi-modal-algos-results}, it is evident from the table that EPO outclassed the rest of the algorithms and its average converges to the global optimum solution with the least standard deviation. It is because of the efficient and vast exploration of the algorithm that then transformed into exploitation when it reaches to the region of optimum solution. As mentioned earlier, multi-modal functions have number of local optimum solution with only one global solution, the results we obtained showed that EPO efficiently avoid the local optima and converges to the global solution only.\\
\section{Analysis of the EPO algorithm} \label{section4}
In Section \eqref{section2} we discussed in detail the performance of both the EPO algorithms we also discuss how EPO (mod.) is more powerful than EPO. In Section \eqref{section3} we did a comprehensive, competitive analysis of the both EPO algorithms with other well-known algorithms which includes; ant-lion (ALO), dragon-fly, particle swarm {\it etc.} From the detail analysis we concluded that in both uni and multi modal functions EPO outperform the testing algorithms and very efficiently achieve the required optimal point.\par
\begin{figure*}[]
    \centering
    \includegraphics[width=1\linewidth]{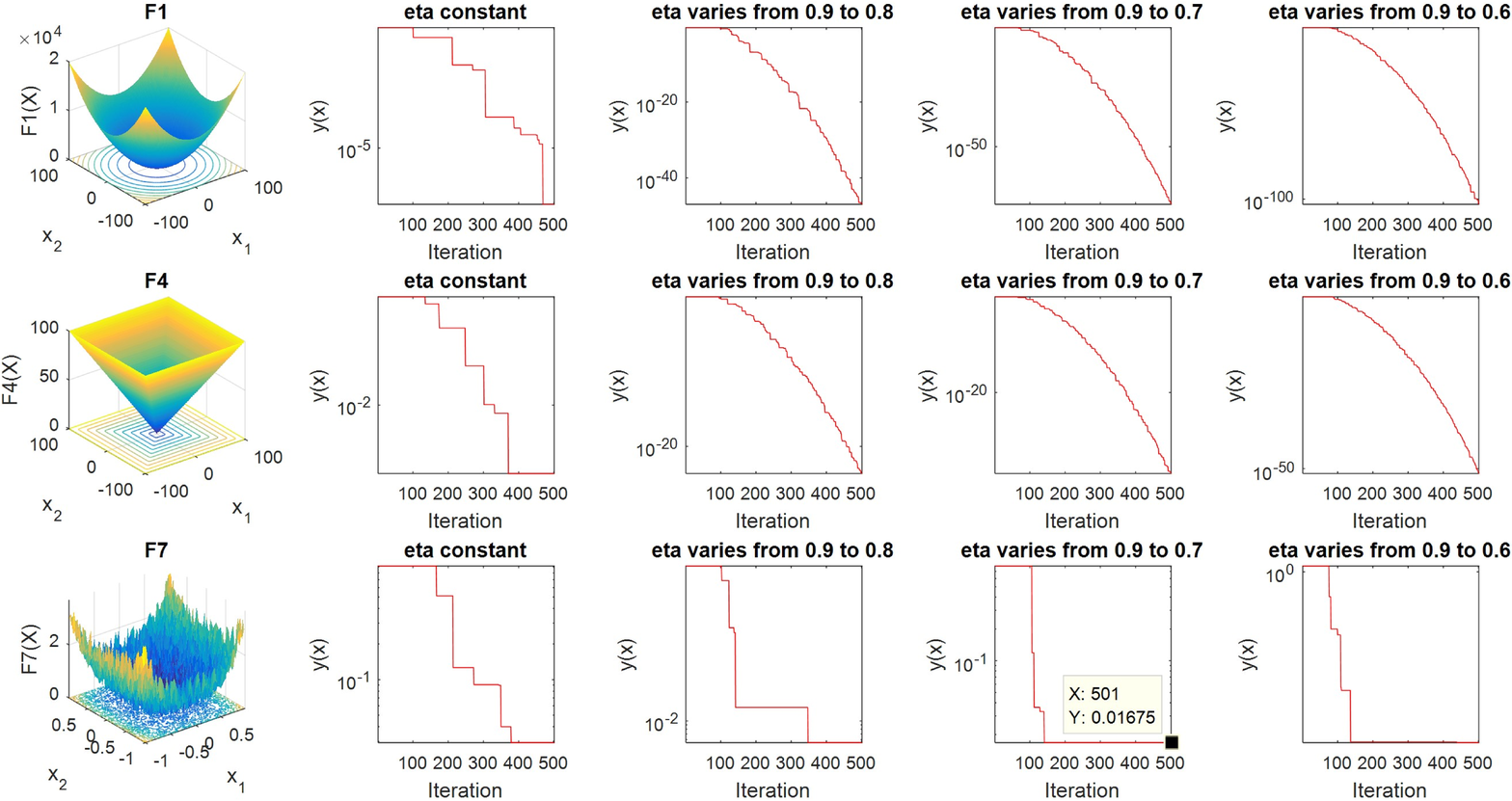}
    \caption{Search space of uni-modal functions, convergence with constant $eta$, convergence when $eta\ varies\ from\ 0.9\ to\ 0.8$, convergence when $eta\ varies\ from\ 0.9\ to\ 0.7$,  convergence when $eta\ varies\ from\ 0.9\ to\ 0.6$}
    \label{analysis_1}
\end{figure*}

\begin{figure*}[]
    \centering
    \includegraphics[width=1\linewidth]{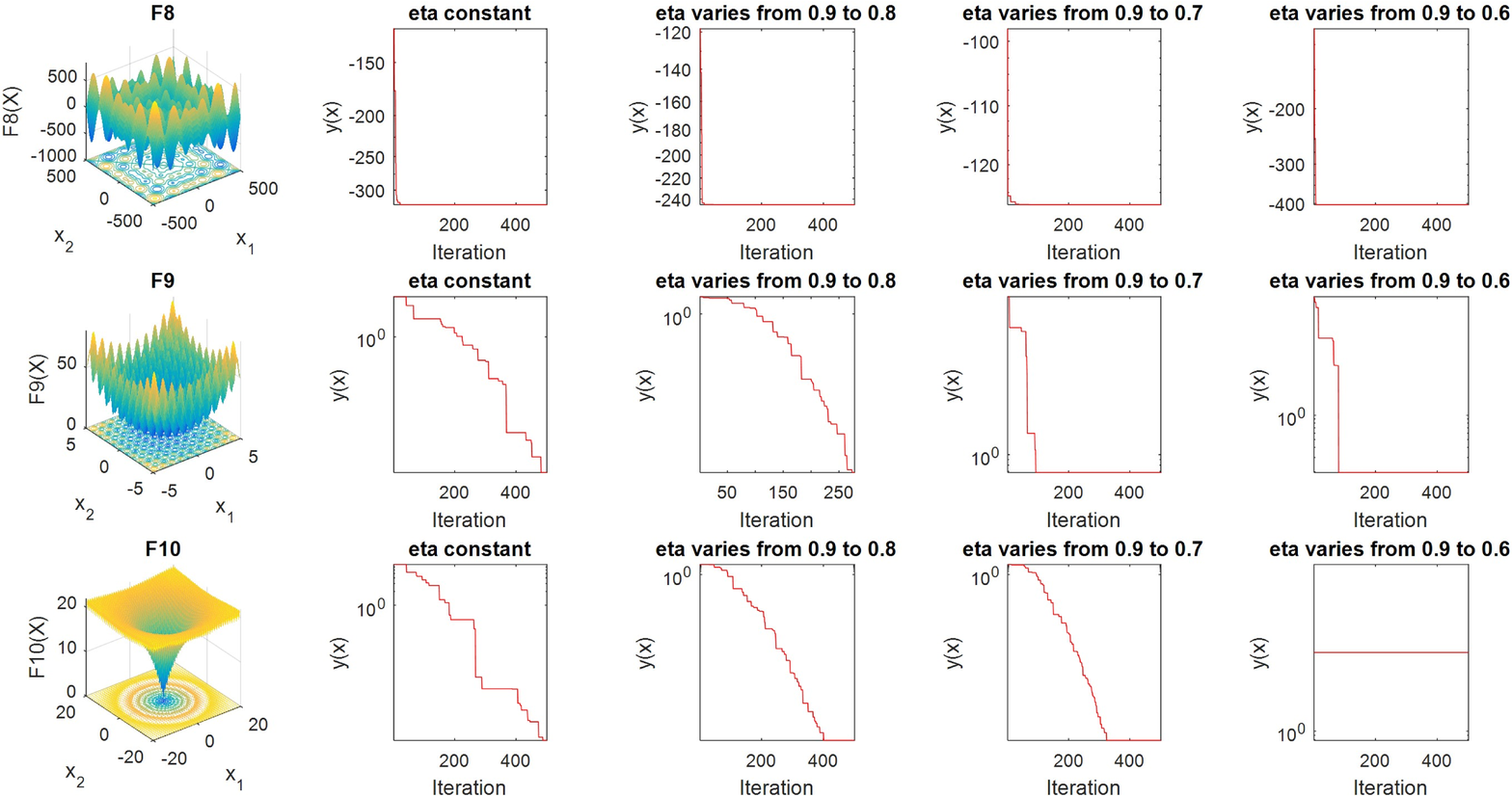}
    \caption{Search space of multi-modal functions, convergence with constant $eta$, convergence when $eta\ varies\ from\ 0.9\ to\ 0.8$, convergence when $eta\ varies\ from\ 0.9\ to\ 0.7$, convergence when $eta\ varies\ from\ 0.9\ to\ 0.6$}
    \label{analysis_2}
\end{figure*}
In this section, we will do the detail analysis of the EPO algorithm, we will manipulate the controlling unit of the algorithm and will find out how it effects the output. Before starting, the parameters of the algorithm particularly for this practice are; $l_scale = 100, res = 0.05, t_{s(iteration)} = 500, particles = 30, and\ dimension = 2$. As explained earlier in modified version of the algorithm we manipulated the value of $eta$ using Equ. \eqref{eta.mod}. The value of $eta$ in fact faster the transformation from exploration to exploitation of our algorithm making it more efficient. To explore the effect of $eta$ on the working of our algorithm we choose three functions; $F1, F4, and\ F7$ from the uni-modal function list \ref{uni_EPO} and run the algorithm with three four different values of the $eta$ i.e. 0.9, 0.9-0.8, 0.9-0.7, and 0.9-0.6.\par
The obtained results are shown in Fig. \eqref{analysis_1}.
It is evident from the figure that as the value of $eta_{min}$ decreases, in other words as the difference between $eta_{min}$ and $eta_{max}$ increases, the test functions converge to their optimal value very fast, and more efficiently. For example, in case of $F1$, if we run the algorithm with $eta\ constant$, after $500$ iterations the function ended up at around $10^{-15}$, same results were obtained within $150$ iterations when $eta\ varies\ from\ 0.9\ to\ 0.8$, same trend is followed further.
All the three functions behave alike but if we meticulously observe the trend in $F7$, we can notice that there is a slight variation and the trend is not as same as in previous cases.
There is a bit of irregularity when we move from $"eta\ varies\ from\ 0.9\ to\ 0.8"$ to $"eta\ varies\ from\ 0.9\ to\ 0.7"$ and further $"eta\ varies\ from\ 0.9\ to\ 0.6"$.
From the first two functions, $F1$ and $F4$, it looks like that the algorithms become more efficient as the range of $eta$ increases.
But, $F7$ breaks this trend and showed that is not the case always.
Namely, as the complexity of a goal function increases we will find certain anomalies in the algorithm functioning with respect to the range of $eta$ Equ. \eqref{eta.equ}.
From the testing of uni-modal functions we concluded two things:

\begin{itemize}
    \item For simple functions, as the range of $eta$ increases the function will faster converges to the optimal point.
    \item For complex functions, as the range of $eta$ increases, we will find anomalies in the system, and there we need to tune the value of $eta$ to achieve the optimal point.
\end{itemize}

We further extended the analysis to the multi-modal problems which are presented in Table \eqref{multi_EPO}.
The generated results are shown in Fig. \eqref{analysis_2}, it proves our above mentioned second point that as the complexity of a test function increases the algorithm does not sustained the general trend of $eta$, and as the range of $eta$ increase the convergence of the functions get affected.
Since multi-modal functions are more complex in nature than uni-modal (because they have number of local-minima and one global minima),
it is not wise to accelerate the transformation from the exploration to exploitation phase, as there is a good chance to avoid the global optimal solution.
It is observed, in that case we need to tune the range of $eta$ to obtain the optimality.\par
Function $F8$ shows a complex search space, out of all the tested functions it is the only one whose optimal solution is dependent on the number of dimensions employed in the algorithm. If we notice, when we benchmarked the EPO algorithm, the optimal solution obtained for this function was different from the one we obtained here. This is because there the dimension was $30$ and here it is $2$. $F1$ shows a fluctuation in the results obtained using different ranges of $eta$, despite the fluctuation if we look at the graph we can clearly see that $std$ of this function is around $370$ considering the different ranges of the $eta$. Similarly, $F9$ also shows the same trend, the common thing in both function was that they were close to their optimal solution despite the fluctuations. The $F10$ shows a very different trend, for the first three cases of the $eta$ it shows more or less the same trend but for $"eta\ varies\ from\ 0.9\ to\ 0.6"$ it shows a very different trend almost a constant line and not closer to the optimal solution as compare to other cases.\par
The other controlling parameters that are involved in the algorithm are; $l_{scale}$ and $res$ both are directly or indirectly related with $eta$. $l_{scale}$ represents the area of search space that our algorithm ill explores, its value should neither be too large nor to small in both cases it may not be able to track the optimum point accurately. $res$ is also a controlling variable because if its value is higher than $l_{scale}$ than we will have $eta>1$ which means that on each iteration $l_{scale}$ will become greater and greater, so it will be unable for the algorithm to track the optimal point.\par
We can further tune the results with the help of $t_{s}$ and the number of runs.
Here, $t_{s}$ represents the number of iterations to obtain more stable output it is better to have it in triple digits, but it varies from function to function, so we can set it accordingly.
After running the algorithm number of times e.g., $10$ to $20$ times, obtain the average $avg$ and standard deviation $std$ of the results, this will further produce the stable solutions.

\begin{figure*}[]
    \centering
    \includegraphics[width=0.7\linewidth]{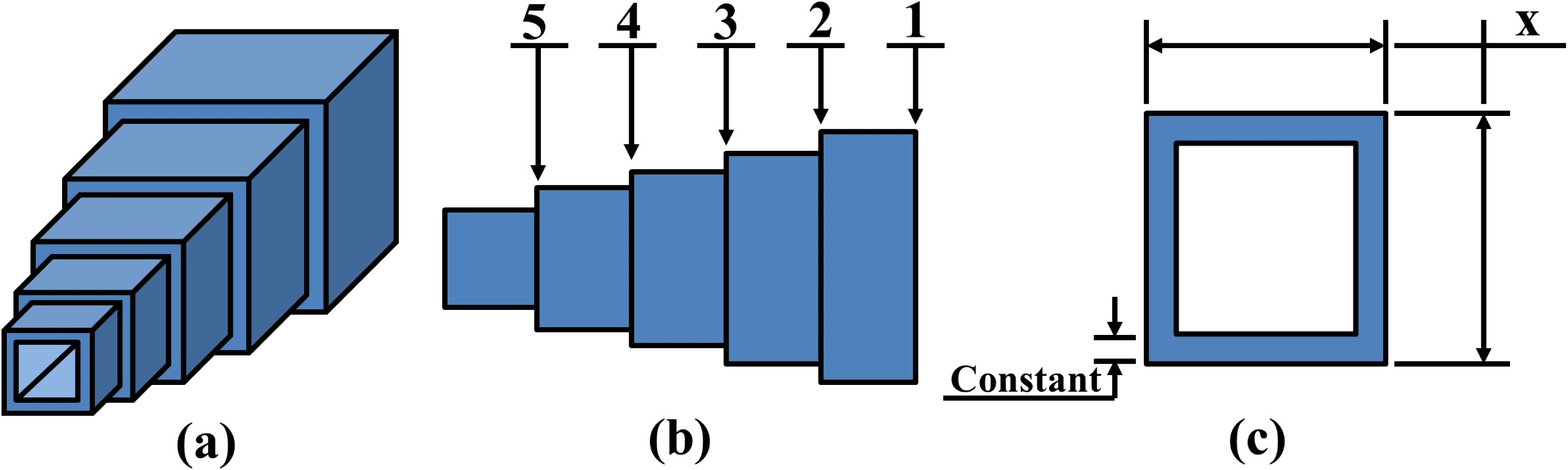}
    \caption{The cantilever beam design problem. (a) is the design of cantilever, (b) shows that all the five blocks of the cantilever are in decreasing size, (c) represents that the thickness of each square block remains constant whereas the length $x$ is the one needs to optimize. }
    \label{fig2}
\end{figure*}

\begin{table*}[h]
    \centering
    \caption{Results of cantilever beam design problem}
    \label{cantilever}
    \begin{tabular} {c c c c c c c c c c}
        \hline
        &Algorithms \hspace{1cm} &  &$x_1$  &$x_2$ &$x_3$ &$x_4$ &$x_5$ &$f_{x}$ &$t_{estd.}(sec)$\\
        \hline \\
        &\multirow{2}{*}{EPO} \hspace{1cm}
        &avg &6.0007&5.2583&4.5051&3.5767&2.1375&13.3683&1.310689\\
        & &std &0.9362E-15&0.9362E-15&0.9362E-15&0.9362E-15&0.4681E-15&1.87E-15
        \\ \\
        &\multirow{2}{*}{EPO (mod.)} \hspace{1cm}
        &avg &6.0123&5.268&4.5392&3.4831&2.1732&13.3665&1.197919\\
        & &std &0.9362E-16&0.000&0.9362E-16&0.9362E-16&0.4681E-16&1.87E-15
        \\ \\
        &\multirow{2}{*}{ALO} \hspace{1cm}
        &avg &6.0338&5.2909&4.4872&3.5071&2.157&13.3667&52.264337\\
        & &std &0.0186&0.0401&0.0305&0.0385&0.0225&8.62E-04
        \\ \\
        &\multirow{2}{*}{DA}\hspace{1cm}
        &avg &6.2205&5.5762&4.4573&3.3765&2.2278&13.6046& 117.31\\
        & &std &0.5335&1.0177&0.2324&0.1874&0.2765&5.00E-01
        \\ \\
        \hline
    \end{tabular}
\end{table*}

\section{Constrained optimization using EPO}\label{section5}

In this section, we will apply the EPO algorithm to some constrained optimization problems and will compare the results with two other algorithms: ALO (ant-lion optimizer) and DA (dragon optimizer).
Each problem has some constraints with it, we will run the problem $10$ times, will $avg$ (average) their optimal coordinates and their respective solutions. We will also calculate the $std$ (standard deviation) and then we will benchmark the performance of algorithms.

\subsection{Cantilever beam design problem}

Cantilever is the structure of five hollow square boxes mounted on each other with the decreasing sizes as shown in Fig. \ref{fig2}a.
Each box has a constant thickness whereas different length $x$, as shown in Fig. \ref{fig2}c.
Here the optimization problem is to minimize the weight of the lever.
There are two constraint for the objective function: variable constraints and vertical displacement constraint \cite{66}.
The mathematical representation of the problem is as follows:
\begin{align*}
&Consider \hspace{0.87cm}  \vv{x} = [x_1, x_2, x_3, x_4, x_5]\\
&Minimize \hspace{0.25cm}  f(\vv{x}) = 0.6224(x_1+x_2+x_3+x_4+x_5)\\
&Subject \hspace{0.1cm}  to\hspace{0.3cm}  g(\vv{x}) = \frac{61}{x_1^3} + \frac{37}{x_2^3} \frac{19}{x_3^3} \frac{7}{x_4^3} \frac{1}{x_5^3}\\
&Variable\hspace{0.1cm} range \hspace{0.3cm}  0.01 \leq x_1,x_2,x_3,x_4,x_5 \leq 100.
\end{align*}

The results obtained after testing the function are shown in Table \eqref{cantilever}.
From the results arranged in the table, it is obvious that both the EPO algorithms have outperformed ALO and DA. Although the $avg$ value of five blocks of both the EPO algorithm was almost same as that of ALO and DA, but least $std$ in both, $std_x$ and $std_{f_x}$.
The $std_x$ and $std_{f_x}$ are far much better than others which mean that the probability of EPO algorithm deviating from its mean value or optimum value in our case is very small.  The time $t_s$ that EPO algorithms took to complete the search is also much lesser than the time corresponding to ALO and DA, this is still a simple engineering problem and AO and DA took a lot of time to complete the algorithm consider a complex problem with several inputs and constraints they will take even more time in that case. Contrary to that, EPO algorithm will save a lot of time and will produce the results more efficiently. This is a practical example that shows the extent to which we can apply this algorithm, it is sufficient enough to handle engineering problem with higher efficiency and accuracy.
\subsection{Three-bar truss design problem}
\begin{figure}[tbh]
    \centering
    \includegraphics[width=1\linewidth]{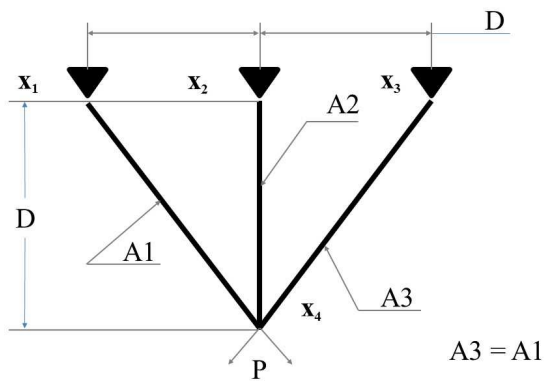}
    \vspace{-1cm};
    \caption{Three-bar truss design problem}
    \label{constrained2}
\end{figure}

The second constrained optimization problem is the design of the three-bar truss to minimize its weight, as shown in Fig. \ref{constrained2}.
The objective function of the problem is very simple, whereas it is highly constrained.
Its structure under goes some severe constraints which include: stress, deflection, and buckling constraints \cite{71}.
The mathematical model of the problem is shown below.

\begin{align*}
&Consider \hspace{1cm}  \vv{x} = [x_1, x_2]\\
&Minimize \hspace{0.4cm}  f(\vv{x}) = (2\sqrt{2}x_1 + x_2)*l\\
&Subject \hspace{0.1cm}  to\hspace{0.3cm}  g_1(\vv{x}) = \frac{\sqrt{2}{x_1}}{\sqrt{2}x_1^2 + 2x_1x_2}P - \sigma \leq 0\\
&\hspace{1.9cm} g_2(\vv{x}) = \frac{x_2}{\sqrt{2}x_1^2 + 2x_1x_2}P - \sigma \leq 0\\
&\hspace{1.9cm} g_3(\vv{x}) = \frac{1}{\sqrt{2}x_2 + x_1}P - \sigma \leq 0\\
&Variable\hspace{0.1cm} range \hspace{0.3cm}  0 \leq x_1,x_2 \leq 1.
\end{align*}

We tested the three-bar truss design in doing so we employed four testing algorithms: eagle perching optimization (EPO),
modified eagle perching optimization (EPO mod.), ant-lion optimization, and dragonfly optimization (DA).
We performed 10 runs on this problem and obtained the $avg$ (average), $std$ (standard deviation), and total time it took $t_{estd.}$.
The results are shown in Table \eqref{three-bar}, which shows the promising results of EPO and EPO (mod.) with the $std$ of range $10^{-15}$ to $10^{-16}$.
The time the considered algorithms took to complete 10 runs is also minimal in case of EPO algorithms around $1.6sec$, whereas ALO and DA are not even comparable with them.

\vspace{0.7cm}
\begin{table}[h]
    \centering
    \caption{Results of three-bar truss design problem}
    \label{three-bar}
    \begin{tabular} {c c c c c c c}
        \hline
        &Function  &  &$x_1$    &$x_2$  &$f_{x}$ &$t_{estd.}$\\
        \hline \\
        &\multirow{2}{*}{EPO}
        &avg &0.79&0.39&2.63&1.22\\
        & &std &0.11E-15&0.058E-15&4.68E-16
        \\ \\
        &\multirow{2}{*}{EPO (mod.)}
        &avg &0.79&0.39&2.63&1.67\\
        & &std &0.11E-15&0.05E-15&0.00
        \\ \\
        &\multirow{2}{*}{ALO}
        &avg &0.78&0.41&2.63&24.20\\
        & &std &0.007&0.01&4.97E-04
        \\ \\
        &\multirow{2}{*}{DA}
        &avg &0.82&0.34&2.66&94.17\\
        & &std &0.06&0.13&6.01E-02
        \\ \\
        \hline
    \end{tabular}
\end{table}

\subsection{Gear train design problem}
\begin{figure}[tbh]
    \centering
    \includegraphics[width=0.7\linewidth]{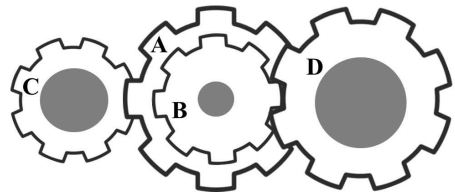}
    \vspace{-1cm};
    \caption{Gear train design problem}
    \label{constrained3}
\end{figure}

This is another problem of constrained optimization, it is related to the making of train gears, shown in Fig. \ref{constrained3} and the optimization problem is to find the optimal number of
tooth for the four gears to minimize the gear ratio.
It is not highly constrained problem, it has only one constrain the range of number of tooth for the gear \cite{76}.
The mathematical formulation of the problem is shown below.

\begin{align*}
&Consider \hspace{1cm}  \vv{x} = [x_1, x_2, x_3, x_4]\\
&Minimize \hspace{0.4cm}  f(\vv{x}) = (\frac{1}{6.931} - \frac{x_3*x_2}{x_1*x_4})^2\\
&Variable\hspace{0.1cm} range \hspace{0.3cm}  12 \leq x_1,x_2,x_,x_4 \leq 60.
\end{align*}
We tested five algorithms and then compared their results with the built-in MATLAB function \texttt{fmincon} which are shown in Table \eqref{gear-train},
we round of the value since we are dealing with the number of tooth for the gears.
From the arranged results, it is evident that EPO algorithms evaluated the results more nearer to the actual solutions.
Especially, EPO (mod.) has not only the more accurate solution but also have the least $std$ (standard deviation) and the evaluation time $t_{estd.}=0.9723$.

\begin{table}[h]
    \centering
    \caption{Results of Gear train design problem}
    \label{gear-train}
    \begin{tabular} {lllllllll}
        \hline
        Function  &  &$x_1$ &$x_2$ &$x_3$ &$x_4$ &$f_{x}$\\
        \hline \\
        \multirow{2}{*}{\texttt{fmincon}}
        &avg &42&16&16&42&3E-12 \\
        &std &N/A&N/A&N/A&N/A&N/A
        \\ \\
        \multirow{2}{*}{EPO}
        &avg &49&15&26&57&3E-18\\
        &std &0.0E-14&0.1E-14&0.0E-14&0.0E-14&8E-34
        \\ \\
        \multirow{2}{*}{EPO (mod.)}
        &avg &42&15&18&44&0.0\\
        &std &0.0E-14&0.0E-14&0.3E-14&0.7E-14&0.0

        \\ \\
        \multirow{2}{*}{ALO}
        &avg &49&19&16&43&2E-12\\
        &std &N/A&N/A&N/A&N/A&N/A
        \\ \\
        \multirow{2}{*}{CS}
        &avg &43&16&19&49&2E-12\\
        &std &N/A&N/A&N/A&N/A&N/A
        \\ \\
        \multirow{2}{*}{MBA}
        &avg &43&16&19&49&2E-12\\
        &std &N/A&N/A&N/A&N/A&N/A
        \\ \\
        \hline
    \end{tabular}
\end{table}

From all the above explained three problems, it is clear that EPO is capable to handle the constrained, real-world problems.
It also proves that its evaluation time is many-fold less than the other test optimization algorithms.
If you further narrow the performance criteria, then EPO (modified)
looks more promising than EPO, because of the ability of EPO (modified) to explore more area in less time.
In other words, the modified EPO has efficient transformation from exploration to exploitation.

\section{Conclusion}\label{section6}

We proposed a novel nature-inspired optimization algorithm, named EPO (Eagle Perching Optimization).
It is based on the behavior of an eagle searching for the highest point in its surrounding to reside on.
We exploited its nature and formulate it into mathematical operators and equations and proposed two algorithms on its basis.
One is simply EPO algorithm with constant rate of decay of search space, whereas the modified version further speeds up the decay and makes it exponential like.
Optimization is all about exploration and exploitation, it makes it more efficient.
Then a detail comparison is done between both the algorithms and later was found more efficient.
We also benchmarked its performance using 10 test functions, both uni-modal and multi-modal and compared the results with another optimization algorithm as well.
Results pointed out that EPO algorithms outperform other algorithms, since they optimize tested functions in lesser number of iterations and with least standard deviation.\par
We did a detail analysis of the algorithm and found that for uni-modal functions the general trend is, as the range of $eta$ increases the efficiency of the algorithm increases. But, this is not the case with complex multi-modal functions, where we need to tune the range of $eta$ but optimum range that we setted for our test functions through hit and trial is from $0.9$ to $0.8$.
We also benchmarked its performance by employing it in solving constrained optimization real world problems, then compared its results not only with other algorithms but also with the built-in function of MATLAB.
We found out that EPO estimated optimal values with higher accuracy and with less amount of time which shows how faster this algorithm works.

%

%






\end{document}